\documentclass[journal,twocolumn]{IEEEtran}
\usepackage{times}
\usepackage{soul}
\usepackage{titlesec}
\usepackage{url}
\usepackage[hidelinks]{hyperref}
\usepackage[utf8]{inputenc}
\usepackage[small]{caption}
\usepackage{graphicx}
\usepackage{amsmath}
\usepackage{amsthm}
\usepackage{booktabs}
\usepackage{multirow}
\usepackage{algorithm}
\usepackage{algorithmic}
\usepackage{comment}
\usepackage{subcaption}
\usepackage[switch]{lineno}

\newcommand\tab[1][.15cm]{\hspace*{#1}}

\ifCLASSINFOpdf
\else
\fi

\begin{document}
\title{Analyzing and Mitigating Bias for Vulnerable Classes: Towards Balanced Representation in Dataset}

\author{Dewant~Katare$^{\mathsection}$,
    David~Solans~Noguero$^\dagger$,
    Souneil~Park$^{\dagger}$,
    Nicolas~Kourtellis$\dagger$,
    Marijn~Janssen$^{\mathsection}$,
    Aaron~Yi~Ding$^\mathsection$,

    $^\mathsection$Delft University of Technology,
    $^\dagger$ \textit{Telef\'onica Research}
\thanks{Corresponding e-mails: d.katare@tudelft.nl}}

\maketitle

\begin{abstract}

The accuracy and fairness of perception systems in autonomous driving are essential, especially for vulnerable road users such as cyclists, pedestrians, and motorcyclists who face significant risks in urban driving environments. While mainstream research primarily enhances class performance metrics, the hidden traits of bias inheritance in the AI models, class imbalances and disparities within the datasets are often overlooked. Our research addresses these issues by investigating class imbalances among vulnerable road users, with a focus on analyzing class distribution, evaluating performance, and assessing bias impact. Utilizing popular CNN models and Vision Transformers (ViTs) with the nuScenes dataset, our performance evaluation indicates detection disparities for underrepresented classes. Compared to related work, we focus on metric-specific and Cost-Sensitive learning for model optimization and bias mitigation, which includes data augmentation and resampling. Using the proposed mitigation approaches, we see improvement in IoU(\%) and NDS(\%) metrics from 71.3 to 75.6 and 80.6 to 83.7 for the CNN model. Similarly, for ViT, we observe improvement in IoU and NDS metrics from 74.9 to 79.2 and 83.8 to 87.1. This research contributes to developing reliable models while enhancing inclusiveness for minority classes in datasets.

\end{abstract}
\IEEEpeerreviewmaketitle

\section{Introduction}

\begin{figure}
\centering
\includegraphics[scale=0.25]{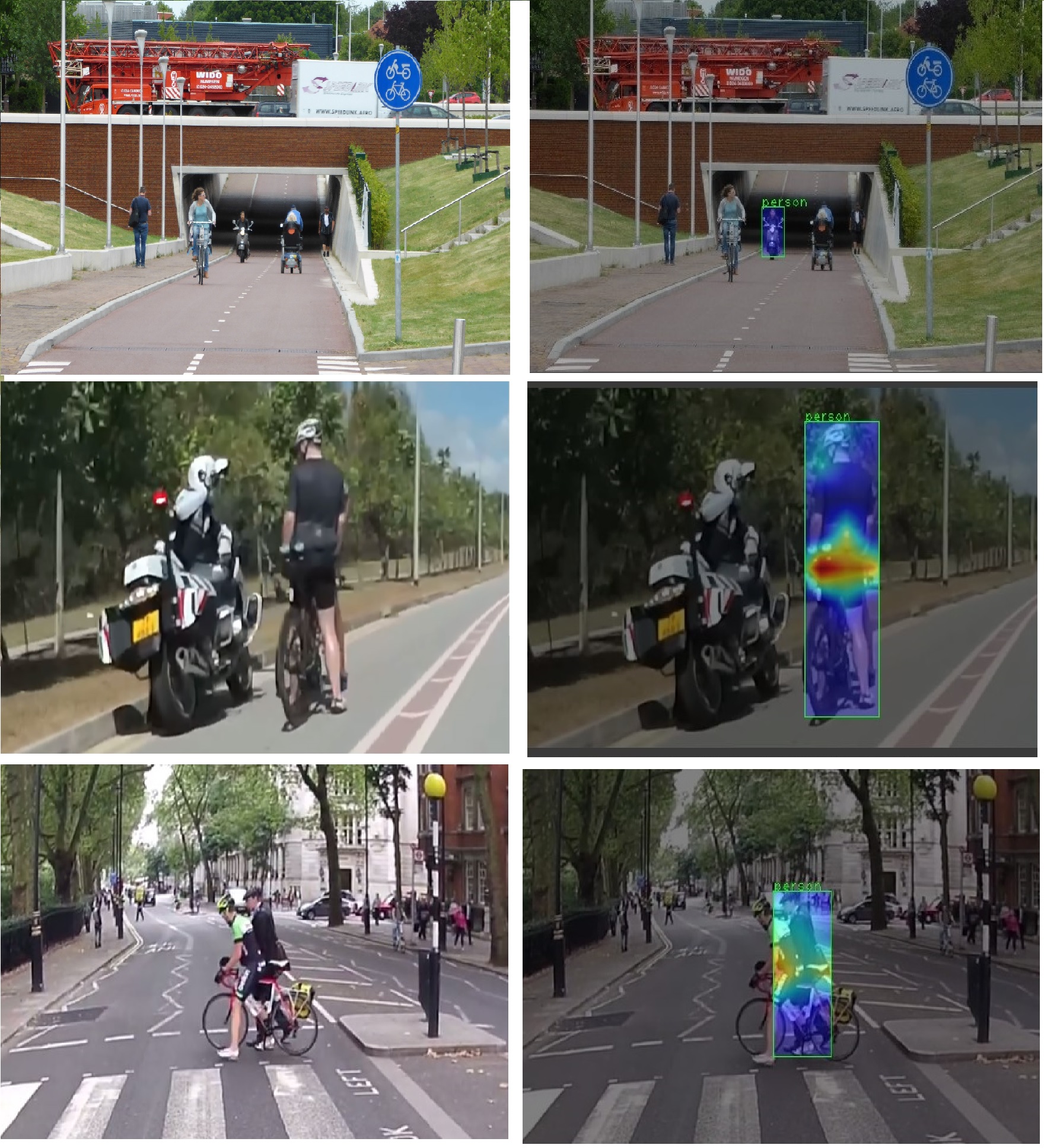}
\caption{Three samples that show testing results when a single class is predominantly present during model training.\\}
\label{bias-case}
\end{figure}

Autonomous driving is dependent on several sensory and AI technologies. The benefits of deploying autonomous vehicles (AVs) have been associated with traffic optimization and safe manoeuvre. However, the fundamentals of such technologies highly rely on the accuracy and fairness of their perception systems \cite{hong2021ai, katare2022bias}. These systems, particularly those involving object detection, are critical for identifying various objects or classes on the road, such as pedestrians, cyclists, motorcyclists, and mobility-impaired persons \cite{fabbrizzi2022survey, wilson2019predictive, katare2023energy}. Recent studies have highlighted the existence of biases in vision datasets, including autonomous driving datasets, which could lead to discriminatory treatment of certain road users~\cite{lee2023resolving, siegel2021morals}. Within this focus, our study discusses two prominent datasets in this domain, nuScenes \cite{nuscenes} and Waymo \cite {waymo_2020}, aiming to address class imbalances and possibilities of biases inheritance in a model when trained with class disparities. This paper covers the representation of vulnerable classes in object detection models’ accuracy and fairness through class distribution analysis, model performance evaluation, and bias impact assessment. Class imbalance refers to the disproportionate representation of different categories or classes within a dataset \cite{saini2021bag, buolamwini2018gender}. Within the context, this often manifests as an over-representation of type of class such as vehicles, while pedestrians, cyclists, motorcyclists, and especially mobility-impaired individuals are classes that can remain underrepresented \cite{lee2023resolving}. This imbalance can lead to perception systems that are less adept at recognizing minor yet crucial classes \cite{mazumder2024dataperf, lee2023resolving, alibrahim2021hyperparameter}. Studies have shown that this imbalance can result in the inheritance of bias in AI models from the datasets, where models preferentially detect overrepresented classes \cite{wang2022revise}. Figure~\ref{bias-case} shows a heatmap from our tests, where images on the left are input images and gradients flow for a class is shown on the right. The consequences of such an imbalance in the dataset cannot be overlooked, as evidenced by incidents like the Uber self-driving car collision \cite{stanton2019models}. We emphasize the need for balanced, diverse and representative datasets that include a wide range of scenarios, reflecting the complexity of real-world conditions. In this dataset imbalance context, our paper explores:
\begin{itemize}
    \item How do the representation and disparities of classes in datasets impact the accuracy of AI models?

    \item How do biases in AI systems for detecting under-represented classes manifest in visual application, and what implications do they have on diverse models?
\end{itemize}

Compared to previous works, which are dependent on the analysis of ``F-1 Score, False Positives(FP) and False Negatives(FN)" for bias impact assessment tasks, our research explores sensitivity and selectivity Score for NNs and Layer-wise relevance propagation for vision transformer models within a framework for class-specific learning/evaluation process to identify bias inheritance. The framework then includes bias assessment using FP, FN, and model re-calibration process to ensure model fairness and effectiveness if a model is retrained on new classes/sub-samples. We also highlight critical gaps in current vision datasets used in autonomous driving through metric-specific learning to develop more equitable and reliable systems. The aim is to contribute to generalising AI algorithms and respective datasets.


\section{Background and Related Work}

Perception technology in autonomous driving has evolved through extensive research and development, with milestones achieved in sensor technology, algorithmic processing, and machine learning models. Early systems relied heavily on rule-based algorithms and limited sensor input, but the advancements in deep learning and improvements in LiDAR, radar, and camera technologies have provided next-generation solutions \cite{katare2023survey}. Despite these advancements, the challenge of developing perception systems that can reliably interpret diverse and unpredictable road scenarios remains an important focus. Datasets play a crucial role in training and validating autonomous driving systems. Datasets like KITTI \cite{geiger2013vision} and Cityscapes \cite{cordts2016cityscapes} have provided foundations for advancing the field. In recent years, datasets such as nuScenes and Waymo have offered more diverse and complex model training and testing environments. These datasets include various annotations for vehicles, pedestrians, cyclists, and other entities, providing a rich foundation for developing and evaluating perception algorithms.

\textbf{Class Imbalance:} Efforts to address class imbalance and bias in autonomous driving datasets have included techniques like data augmentation, re-sampling, and advanced algorithmic approaches like cost-sensitive learning \cite{haixiang2017learning, buolamwini2018gender, mazumder2024dataperf}. Research in this area seeks to balance the representation of classes in training data \cite{torralba2011unbiased} and develop fair and accurate models across all categories \cite{wang2022revise} \cite{wilson2019predictive}. In paper\cite{hahner2021fog}, the author highlights the use of scenario simulations to test and improve model performance in detecting imbalanced classes. The ethical implications of biased autonomous driving systems are significant. There is a growing discourse on the need for equitable AI systems that ensure the safety and fairness of all road users \cite{shariff2021safe}. Furthermore, regulatory bodies are increasingly focusing on how these technologies adhere to safety standards, such as their ability to detect and respond to diverse and representative~\cite{chasalow2021representativeness} scenarios. Targeting imbalanced representation, an approach that combines visual codebook generation with deep features and a non-linear $Chi^2$ SVM classifier, is proposed in \cite{saini2021bag}. This method tackles the issue of imbalanced datasets, where algorithms often fail to detect minority classes. The approach extracts low-level deep features using transfer learning with the ResNet-50 pre-trained model and k-means clustering to create a visual codebook. Each image is then represented as a Bag-of-Visual-Words (BOVW), derived from the histogram of visual words in the vocabulary. The $Chi^2$ SVM classifier is used for classifying these features, demonstrating optimal performance in empirical analysis. This method shows superior accuracy, F1-score, and AUC metrics results compared to state-of-the-art methods, as validated on two challenging datasets: Graz-02 and TF-Flowers \cite{saini2021bag}. 

With a focus on training methods to address the challenges of learning from imbalanced data, a new loss function that mitigates the impact of samples leading to overfitted decision boundaries is proposed in \cite{Park_2021_ICCV}. This loss function has been shown to enhance the performance of various imbalance learning methods. The proposed approach is versatile and can be integrated with existing resampling, meta-learning, and cost-sensitive learning methods to tackle class imbalance problems. The paper validates this method with experiments on multiple benchmark datasets, demonstrating its superiority over state-of-the-art cost-sensitive loss methods. Additionally, the authors have made their code publicly available, facilitating further research and application \cite{Park_2021_ICCV}. An approach for predicting driver behaviour, which is crucial for safely integrating autonomous vehicles into human-dominated traffic, is presented in \cite{driving-bias}. The proposed method addresses the shortcomings of existing predictive models, which either lack transparency (deep neural networks) or are not expressive enough (rule-based models). The authors introduce a model that embeds the Intelligent Driver Model (IDM), a rule-based approach, into deep neural networks. This hybrid model combines the long-term coherence and interpretability of rule-based models with the expressiveness of deep learning, aiming to accurately predict driver behaviour in complex scenarios like merging. The method is an attempt to bridge the gap between two modelling paradigms. It enhances the interpretability of neural network predictions while maintaining accuracy, a critical factor for real-time decision-making in autonomous driving. The model's transparency is particularly beneficial for debugging and understanding its predictions. 

\textbf{Model learning and Representation:} 
The paper by Wang et al. \cite{wang2021deep} introduces a deep attention-based method for imbalanced image classification (DAIIC), utilizing an attention mechanism within a logistic regression framework to prioritize minority classes in prediction and feature representation. It automatically determines misclassification costs to aid discriminative feature learning. Robust across different networks and datasets, the DAIIC method has proven effective, surpassing several benchmarks in single-label and multi-label contexts \cite{wang2021deep}. The background and related work in autonomous driving technologies, datasets, and the challenges of class imbalance and bias underscore the importance of this research area. Our study builds upon these foundations, aiming to contribute to the development of more equitable and reliable autonomous driving systems. By addressing class imbalance and bias in key datasets, we seek to enhance the safety and fairness of these technologies for all road users.

\begin{figure}[!h]
\centering
\includegraphics[scale=0.025]{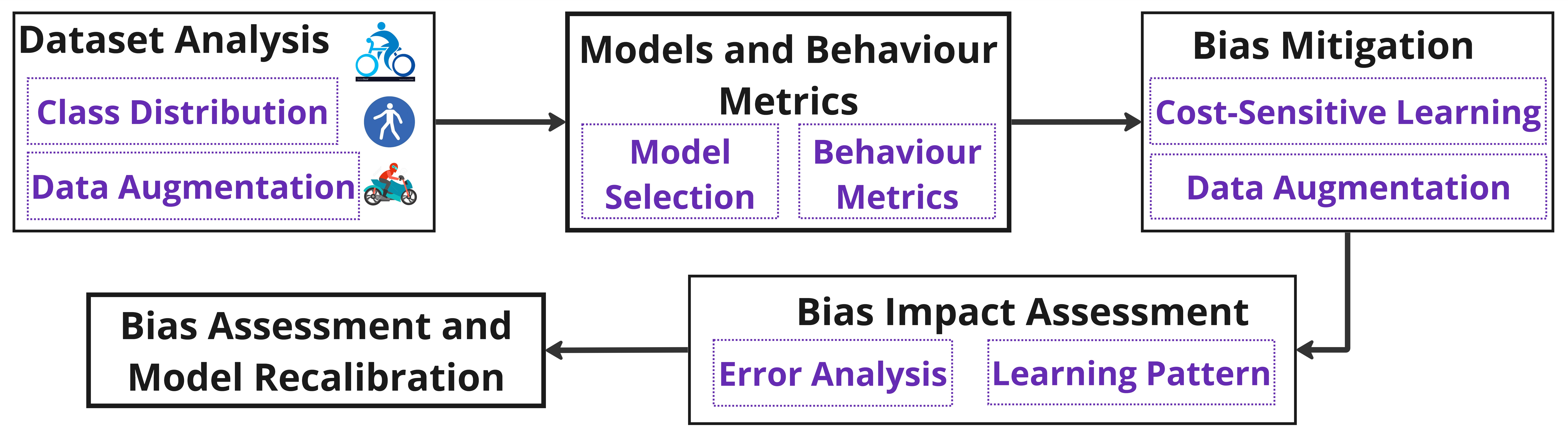}
\caption{Proposed methodology for analyzing and mitigating biases}
\label{method-bias}
\end{figure}

\section{Methodology}

When an imbalanced dataset $D$ with an unjustified distribution of classes ${C_1, C_2,....., C_n}$ is used to train machine learning models, specifically neural networks and Vision Transformers (ViTs), the models are likely to inherit biases because of class representation in the dataset. These biases may arise due to imbalanced class representation and a predominance of normal conditions as compared to challenging real-world scenarios. For neural networks, we anticipate biases to be reflected in neurons' sensitivity and selectivity scores, while for vision transformers, biases may manifest in the attention allocation. Therefore, we propose a methodology consisting of statistical analysis for analyzing and mitigating biases using dataset analysis, bias impact assessment, bias mitigation, metrics behaviour analysis, and bias assessment and model re-calibration, as shown in Figure~\ref{method-bias}. The following subsections detail and discuss the concepts and terminologies used in our methods and tests.

\subsection{Dataset and Classes}
In this paper, we use nuScenes as the case study, focusing on the Pedestrian, Cyclist, and Motorcyclist classes. However, the approach can be tested and validated on other driving datasets, e.g., the Waymo dataset \cite{waymo_2020}, where the samples and classes can be varied. 

\textbf {A) Class Distribution Analysis:} We perform statistical analysis on the frequency of the specified classes in the nuScenes and Waymo datasets to identify class imbalances. The datasets are scrutinized to understand the context (urban versus rural settings, diverse weather and lighting conditions) in which these classes are represented, highlighting any underrepresented scenarios.

{\textbf{Pedestrian Class:}} nuScenes dataset includes a major proportion of pedestrian annotations, with 149,921 instances labelled as ``human.pedestrian.adult" accounting for 21.61{\%} of all annotations. This substantial representation highlights the importance of pedestrian detection in autonomous driving systems. However, other pedestrian sub-categories like children (1,934 annotations, 0.28{\%}), construction workers (13,582 annotations, 1.96{\%}), and individuals with personal mobility devices (2,281 annotations, 0.33{\%}) are less represented. These differences shows gaps in dataset and subsamples diversity.

\noindent
{\textbf{Cyclist Class:}} It is categorized as ``vehicle.bicycle"  and has around 17,060 annotations, comprising 2.46{\%} of the total dataset. This relatively lower representation compared to pedestrians impacts the class weight calculation and, therefore, the model's ability to accurately detect cyclists, a safety concern in urban driving scenarios.

{\textbf{Motorcyclist Class:}} It is annotated as ``vehicle.motorcycle" and has around 16,779 annotations, making up 2.42{\%} of the dataset. This class is slightly underrepresented compared to pedestrians but is on par with cyclists. The lesser representation of cyclists and motorcyclists indicates an area for improvement in class balance, especially considering the different dynamics and associated risks.

The standard metrics used for model performance evaluation in nuScenes are \textit{mean average Precision} (mAP), \textit{Intersection over Union} (IoU), and \textit{nuScenes Detection Score} (NDS). mAP and IoU are well-known metrics used in 3d object detection and segmentation tasks \cite{nuscenes, katare2023energy}, and are calculated using the fundamentals of precision, recall, area of overlap and intersection over union respectively. Similarly, NDS combines several metrics, including the mAP, to provide a comprehensive performance score for the object detection models. The NDS metric is calculated as:

\begin{equation}
NDS = \frac{1}{10} [5 \cdot mAP + \sum_{mTP \in TP} (1 - \min(1, mTP))]
\end{equation}

\textbf {B) Data Augmentation/Re-sampling:} Given the class representations, particularly for the less represented sub-categories of pedestrians and overall lower representation of cyclists and motorcyclists, following strategies can be used for model training and evaluation.

\textbf{Resampling Technique:} Addressing class imbalances in the datasets for underrepresented classes, like cyclists and motorcyclists, is crucial for unbiased training in autonomous driving models. We apply both ``Random Oversampling" and ``Undersampling " from resampling techniques". Random Oversampling duplicates instances from minority classes, ensuring models have sufficient data to learn from these critical yet less frequent road users. Conversely, Undersampling reduces the predominance of over-represented classes, like pedestrians, to prevent bias inheritance in models. Both methods are carefully balanced to maintain data diversity \cite{nie2021resampling}. 

\noindent
\textbf{Data Augmentation:} Another key approach is the use of geometric transformations, such as rotating and flipping images. This technique represents various class orientations, which is crucial for autonomous driving where objects are encountered from multiple directions. These transformations prevent the model from developing orientation-specific biases and enhance its ability to adapt to diverse real-world scenarios. Additionally, this approach contributes to increasing the dataset's diversity and size, thereby improving the model's robustness and generalization capabilities for more accurate and reliable perception tasks \cite{rubaidi2022fraud}.

\noindent
\textbf{Combination of Strategies:} For varied training, testing, and to increase the samples/sub-classes diversity, we combine resampling and data augmentation to balance the training dataset. This approach integrates the strengths of both Resampling (Random undersampling/Oversampling) and Transformations, such as rotating and flipping images, to optimize the representation and variability of data.

\subsection{Models and Behavioural Metrics}
We strategically use popular models such as ResNet18, SqueezeNet from the CNN and ViT from the transformer family for evaluation. The selection of these models is intended to provide an understanding of how different architectures perform on biased datasets and to compare their ability to generalize across various classes, especially after the implementation of bias mitigation. Using these models, we further gain insights across different neural network architectures, ensuring robustness and reliability in our findings.

\noindent
\textbf{Model Selection:} ResNet18 \cite{li2019fully} is a well-known architecture and performs fairly in perception tasks due to its ability to capture spatial hierarchies, but it can show a tendency of bias towards common textures and patterns. SqueezeNet \cite{iandola2016squeezenet}, is a compact model and focuses on global image features but risks inheriting biases from non-diverse datasets \cite{katare2022bias}. Vision Transformers (ViTs) process images by analyzing sequences of patches, adeptly focusing on relevant image parts due to their attention mechanisms \cite{dosovitskiy2020image}. While ViTs are proficient at capturing dependencies, they may struggle to adequately focus on underrepresented classes/unique scenarios. Each model’s inherent strengths and learning mechanisms influence classification and learning, underscoring the need for tailored bias detection and mitigation strategies.

\noindent
\textbf{Behaviour Metrics:} Sensitivity and selectivity scores of neurons are used to evaluate biases in models like ResNet and SqueezeNet. Sensitivity scores measure a neuron's response to changes in input from specific classes, indicating potential biases. For example, a ResNet neuron more sensitive to vehicles than cyclists might show a vehicle-detection bias. Selectivity scores, on the other hand, assess the specificity of a neuron’s response to a class, with high selectivity indicating specialized recognition capabilities. This helps identify whether models effectively differentiate between classes or exhibit biases, such as a score showing higher selectivity for pedestrians over motorcyclists, potentially indicating a feature recognition bias. The formula is described as: 
$$\text{Sensitivity Score} = \partial a / \partial x \text{\;}$$
where \( a \) is activation of neuron, and \( x \) is the input feature.
$$\text{Selectivity Score} = \frac{a_c - a_{avg}}{\max(a_c, a_{avg})} \text{\;}$$

where \( a_c \) is activation of the neuron for the target class and \( a_{avg} \) is the average activation for all classes. 

In Vision Transformers (ViTs), attention map analysis is key for identifying model focus areas and potential decision-making biases. This involves extracting attention weights from each Transformer layer, which indicates the model's focus on different image parts. These heatmaps created using the scaled dot-product attention mechanism, help determine if the model disproportionately focuses on certain features, potentially signaling bias. Attention weights are calculated as $A = \text{softmax}\left(\frac{QK^T}{\sqrt{d_k}}\right)V$, where \( Q \) (Query), \( K \) (Key), and \( V \) (Value) are matrices derived from the input, and \( d_k \) is the dimension of the Key vectors. By analyzing these attention maps across different layers and attention heads, we can identify whether the model disproportionately focuses on certain features or misses critical aspects, potentially indicating bias. 

Layer-wise Relevance Propagation (LRP) offers another approach, starting from the output layer and backpropagating the relevance score of the predicted class through the network. LRP redistributes each layer’s relevance to its inputs, particularly in self-attention layers, where relevance among patches is allocated based on attention weights \cite{ali2022xai}. The relevance for a patch in a layer, highlighting the model's decision-making basis is calculated as:
\begin{equation}
R_j^{(l)} = \sum_{i} A_{ij}^{(l)} R_i^{(l+1)}
\end{equation}
Where \( R_j^{(l)} \) is the relevance of patch \( j \) in layer \( l \), \( A_{ij}^{(l)} \) are the attention weights from patch \( j \) to patch \( i \) in the self-attention mechanism, and \( R_i^{(l+1)} \) is the relevance in the next layer.

\subsection{Bias Impact Assessment}
This section describes model errors (false positives and negatives) and examines learning patterns and bias inheritance, which are crucial for identifying potential biases in model performance.

\textbf{A) Error Analysis:} We focus on false positives (e.g., incorrectly identifying objects as belonging to a target class) and false negatives (e.g., failing to detect cyclists). We further use \textit{Class-specific error} analysis, which includes evaluating how well the model detects pedestrians, cyclists, and motorcyclists under various scenarios. We also examine if there is a tendency to overlook certain classes and how this impacts overall model performance. The correlation between these errors and previously identified biases, such as neurons' sensitivity or selectivity scores, is also explored to understand the underlying causes and guide bias mitigation strategies using \textit{bias correlation}.

\textbf{B) Learning Patterns and Bias Inheritance:} Different model architectures lead to varied learning behaviours and potential biases. CNNs like ResNet may focus on textures and local patterns, which could introduce biases if such features are not uniformly present across all classes. This can result in misclassification, especially when distinguishing features are missing. In contrast, Vision Transformers (ViTs) may develop biases based on how attention is distributed across an image.

\noindent
\textbf{Bias Inheritance from Data:} A model's performance can vary depending on the diversity/samples encountered in training. For instance, a model trained predominantly on urban pedestrian images might underperform in rural settings. Understanding this bias inheritance is crucial for ensuring that perception systems are adaptable to diverse real-world conditions and can guide necessary adjustments in data representation and model training.

\noindent
\textbf{Comparative Analysis:} By comparing how different models like ResNet, SqueezeNet, and ViTs respond to the same dataset, we can uncover specific biases inherent in each architecture. This comparative approach helps identify which models are more prone to biases, helping categorise the most suitable model for bias-sensitive applications.

\noindent
\textbf{Interpretation and Visualization:} Layer-wise Relevance Propagation (LRP) provides insights into what drives a model's decisions by highlighting influential input features. For example, if a model consistently focuses on irrelevant background features for decision-making, it indicates a bias that needs correction \cite{montavon2019layer}. Analysis of these scores, using visual tools like heatmaps and attention maps, helps in interpreting these complex models, offering an understanding of their decision-making processes, and adjusting the training process to mitigate identified biases. Overall, this process involves the analysis of learning patterns, data bias inheritance, comparative model analysis, and visualization, which can help discover bias occurrence/inheritance in different model architectures.

\subsection{Bias Mitigation Techniques}
We explore the strategies mentioned below to mitigate biases inherited from sample representation and address these class disparities. These techniques are necessary for addressing the disparities and ensuring that the model's performance is equitable and reliable.

\textbf{A) Cost-Sensitive Learning:} Customized cost-sensitive learning offers a robust approach to mitigating these issues by adjusting the model's loss function according to the representation of each class. The class weights calculation and loss function are discussed below.

\textbf{Class Weights Calculation:} Based on the dataset statistics, we calculate the weights for the pedestrian ${w_p}$, cyclist ${w_c}$, and motorcyclist ${w_m}$ classes using the inverse of their respective representation percentages. This approach assigns higher weights to underrepresented classes, emphasizing their importance during training. The calculated weights are:

\begin{equation}
{w_p} = \frac{1}{21.61{\%}}; \tab {w_c} = \frac{1}{2.46{\%}}; \tab {w_m} = \frac{1}{2.422{\%}}
\end{equation}

These weights are then normalized to ensure they contribute proportionally and maintain stability during training.

\textbf{Loss Function:} We integrate these calculated weights into the model's loss function, utilizing a weighted multi-class cross-entropy loss. For each instance {\textit{i}}, the loss is defined:

$L_i = - w_p y_{ip} \log(p_{ip}) - w_c y_{ic} \log(p_{ic}) - w_m y_{im} \log(p_{im}) \text{\; (4)} $

Here ${w_p}$, ${w_c}$, and ${w_m}$ are the weights for pedestrians, cyclists, and motorcyclists. ${y_{ix}}$ is a binary indicator of whether instance {\textit{i}} belongs to a class {\textit{x}}. ${p_{ix}}$ is the model's predicted probability for instance {\textit{i}} belongs to a class {\textit{x}}. 

\textbf{Dynamic Adjustment Evaluation:} To ensure continual model adaptability we propose dynamically adjusting these weights based on performance metrics. This approach can be tested using a validation set to monitor performance improvements for underrepresented classes and to prevent overfitting.

By using this customized cost-sensitive learning approach, we aim to reduce the biases inherent in the AI model from the imbalanced dataset, ensuring a more equitable and robust model performance across all classes, which is crucial for the reliability and safety of perception applications.

\textbf{B) Data Augmentation Based on Model Analysis:} We use a targeted data augmentation strategy to address biases identified in ViTs through attention map analysis and Layer-wise Relevance Propagation (LRP). This approach specifically addresses under-representation issues in cyclists and motorcyclists within the nuScenes dataset.

\textbf{Attention-Guided Augmentation:}
Insights from attention maps guide this augmentation strategy. For instance, if the model frequently overlooks pedestrians under night or certain lighting conditions, then there is a need for more class samples highlighting these scenarios. Techniques such as zooming or adjusting brightness/contrast are used to emphasize these aspects. Similarly, additional variations of poses or orientations for motorcyclists are incorporated to improve model recognition abilities in these contexts.

\textbf{LRP-Informed Sampling:} Layer-wise Relevance Propagation provides an understanding of which features contribute most to the model's decisions. Using insights from LRP, the dataset can be augmented to enhance the representation of features critical for correct classifications. Augmentation for this case includes scenarios where the model typically misclassifies images, like partially occluded subjects or specific textures and patterns, enhancing the model's ability to differentiate between relevant and misleading features.

\textbf{Implementation:} This strategy involves iterative dataset refinement based on continuous model analysis. By aligning augmentation closely with model performance, we aim to mitigate biases, ensuring balanced and fair model performance. This is critical for perception systems, where the accurate detection of all road users, particularly those underrepresented like cyclists and motorcyclists.
    
\subsection{Bias Assessment and Model Re-calibration}
Effective bias mitigation requires continuous evaluation and adjustment of models. This section describes our approach to bias assessment and model re-calibration.

\textbf{Bias Assessment:} In this step, we use behaviour metrics, specifically sensitivity and selectivity scores for each class, to measure the impact of our mitigation efforts. Comparing these metrics before and after mitigation measures provides insights into changes in model behaviour. Additionally, we analyze error rates, particularly false positives and negatives, to assess improvements in the model's predictive accuracy across different classes. For Vision Transformers, attention map analysis is used to verify if attention distribution is now more balanced across various classes and scenarios.

\textbf{Model Re-calibration:} This involves dynamically adjusting class weights within the loss function, guided by real-time performance metrics. The step ensures the model stays optimized for any shifts in class representation or emerging imbalances. When new class samples or data become available, the model undergoes re-training to stay aligned with the evolving operational context. The last step includes an iterative refinement, monitoring and updating bias mitigation strategies based on the latest performance assessments.

\section{Experimental Evaluation}
This section covers training, testing and evaluation methods. The overall dataset comprises approximately 1.4 million images, with annotations across several classes. Our focus is on the Pedestrian (149,921), Cyclist(17,060), and Motorcyclist (16,779) classes, as it provides a rich source for analyzing biases in object detection.

\textbf{Preprocessing Steps:} Data preprocessing involved normalization of image pixel values to the [0,1] range, aligning annotations to ensure consistency, and resizing images to standard dimensions suitable for input requirements of chosen models.

\textbf{Model Architectures:} The study uses SqueezeNet and ResNet18 from the CNN family alongside Vision Transformer: ViT. SqueezeNet, known for its compressed size and robustness in feature extraction, and ResNet18, recognized for its efficiency in learning from residual connections, are expected to provide insights into traditional CNN performance. ViT, representing the Transformer family, is included to evaluate the effectiveness of its attention-based mechanism in handling class imbalances.

\subsection{Hyperparameter Optimization}
Initial hyperparameters were set as follows: learning rate of 0.001, batch size of 32, and weight decay of 0.0001 for CNN models. For ViT we configured the training with a batch size of 32 and an initial learning rate of 1e-3, following a linear decay to 1e-5. The model was trained for 30 epochs using the Adam optimizer. A dropout rate of 0.1 and a weight decay of 0.03 is applied to avoid overfitting.

\textbf{Optimization Techniques:} Hyperparameter tuning is performed using a grid search approach, systematically varying learning rates, batch sizes, and dropout rates to identify the optimal combination \cite{alibrahim2021hyperparameter}. The optimal set of hyperparameters was chosen based on the highest average scores across these metrics as followed in \cite{nuscenes}, focusing on improvements in detecting classes.

\subsection{Training Process}
Our sample size is small, so CNN models were trained for 50 epochs using an Adam optimizer. A learning rate scheduler was used to reduce the learning rate by 10{\%} every 10 epochs. Regularization techniques such as dropout and data augmentation (rotations and flipping) were used to prevent overfitting \cite{keshari2018learning}.

The dataset is divided into 70{\%}~training, 15{\%}~validation, and 15{\%}~test splits, ensuring a fair representation of scenarios in each set. Training is conducted on a high-performance computing cluster with a combination of NVIDIA Tesla V100 GPUs using Open MPI and PyTorch as the deep learning framework.

\begin{figure}[!ht]
       \centering
  \includegraphics[scale=0.25]{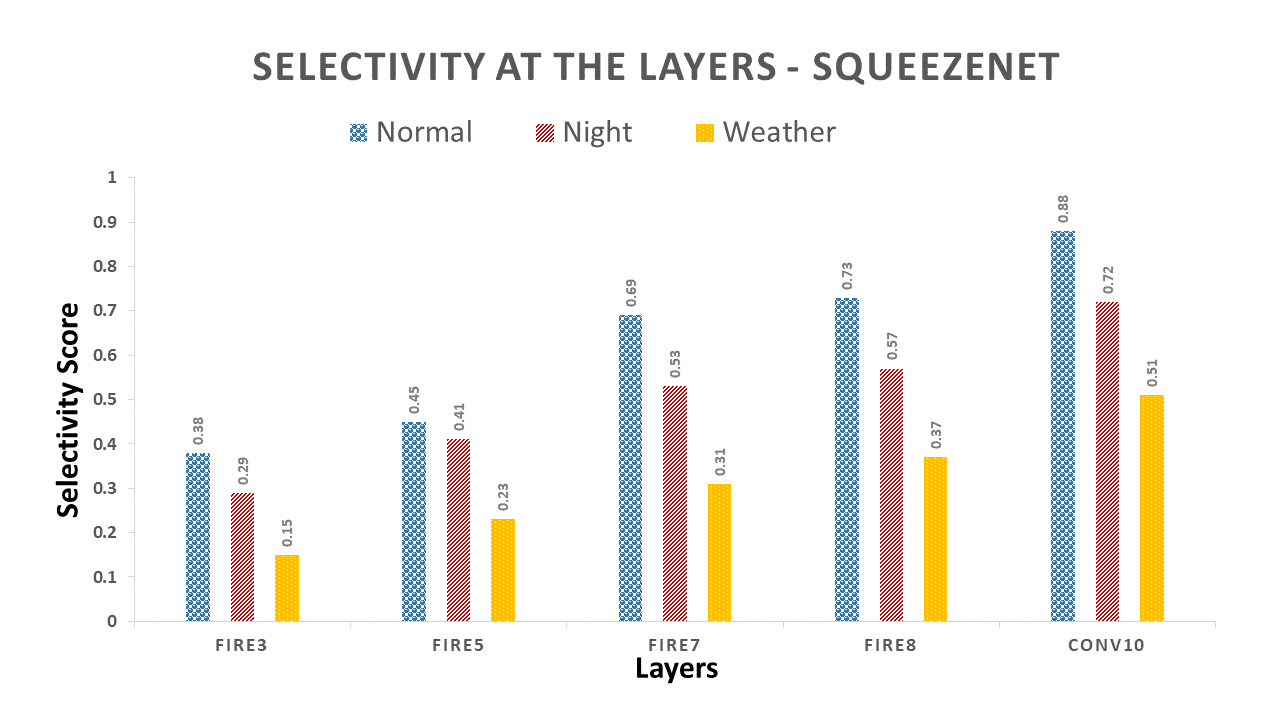}
  \caption{Figure shows layerwise selectivity for the Classes}
  \label{error-rarte}
\end{figure}

\subsection{Model Evaluation}
Models were evaluated using mAP, {\%}IoU and NDS (Intersection over Union, NuScenes Detection Score), alongside sensitivity and selectivity scores for class analysis.

\textbf{Baseline Comparison:} Model performances pre and post-implementation of bias mitigation strategies were compared. Baseline models were trained with respect to data conditions.

\textbf{Class-specific Analysis:} Performance metrics for each class were used to assess improvements in detecting pedestrians, cyclists, and motorcyclists. For analysis, attention is given to false positive or negative rate changes for these classes. The tests for CNN can be performed using layer-wise analysis and by varying data diversity where one class has a dominant presence.

\begin{figure}[!ht]
\centering
  \includegraphics[scale=0.35]{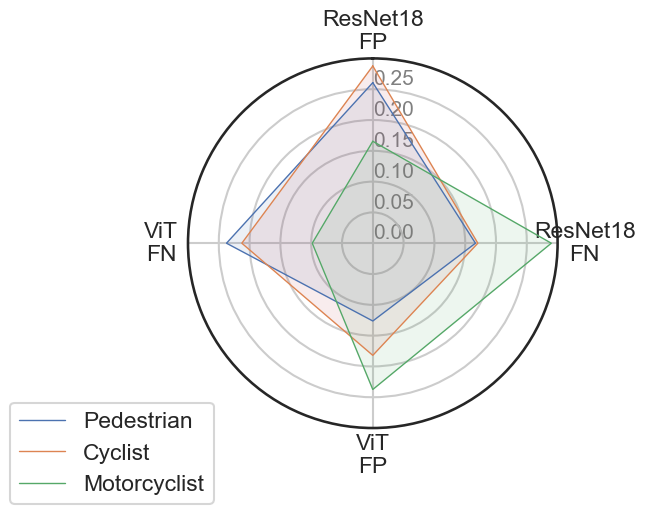}
  \caption{Visualization of class-wise error rate for ResNet and ViT according to False Negatives and False Positives in model validation.\\}
  \label{error-rarte}
\end{figure}

\textbf{Layer-wise Analysis:} This analysis reveals whether specific layers are biased towards particular classes, which provides strategies for bias mitigation. For instance, if early layers are biased towards detecting pedestrians more than cyclists or motorcyclists, there will be a need to adjust the training data with inclusiveness.

\noindent
\textbf{Analysis over Epochs:} Monitoring sensitivity and selectivity over epochs allows you to see if the model improves its ability to recognize underrepresented classes. If sensitivity and selectivity for these classes do not improve, or if they plateau early, this could indicate that the model is not effectively learning features for these classes, suggesting that further intervention is required. A layerwise selectivity for the three classes in different scenarios is shown in Figure~\ref{error-rarte}.


\begin{figure}[!ht]
       \centering
  \includegraphics[scale=0.15]{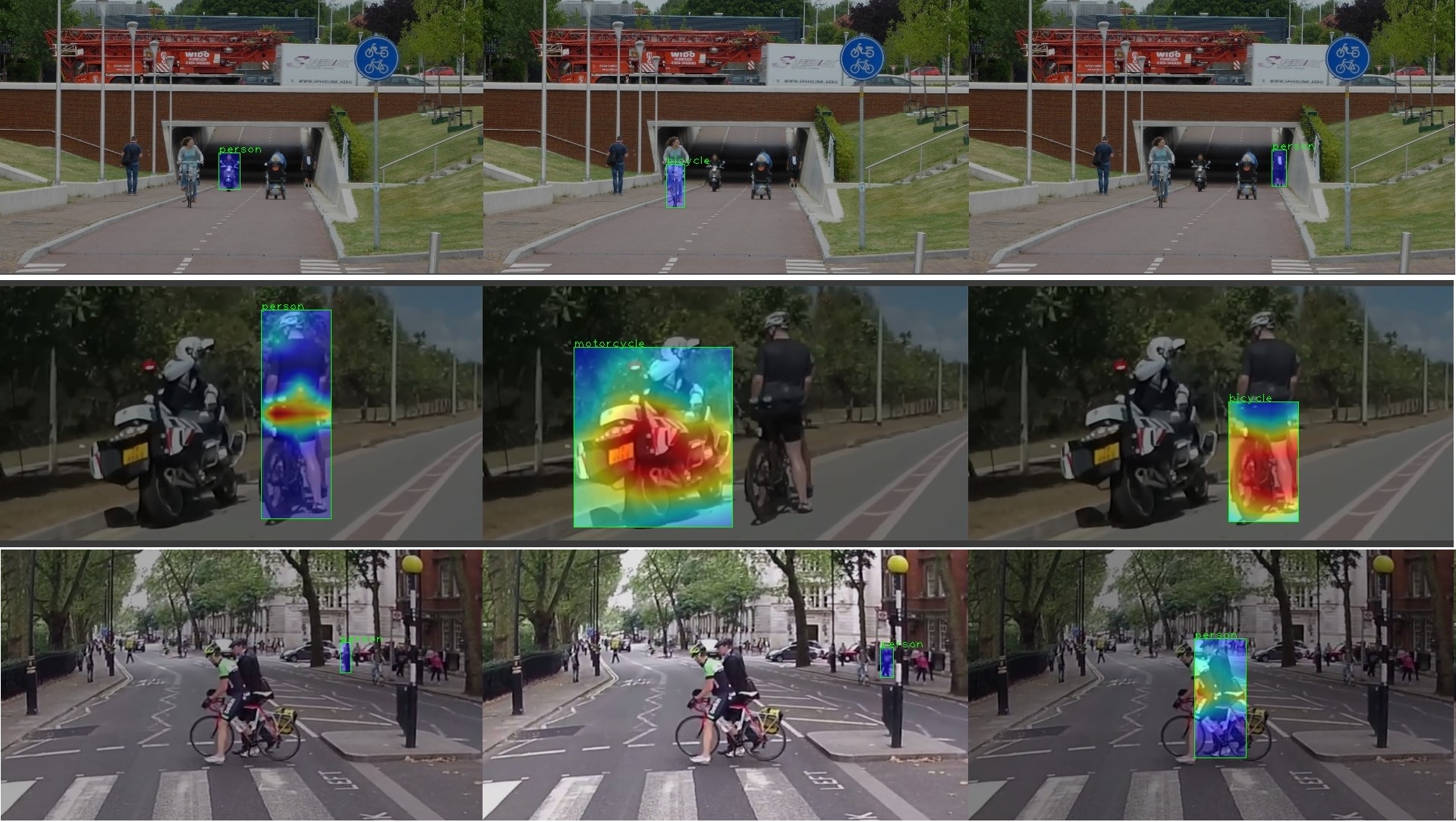}
  \caption{Figure showing heatmap for all three classes.\\}
  \label{class-hm}
\end{figure}

\subsection{Bias Impact Analysis}

\textbf{Error Analysis:} Figure~\ref{error-rarte} shows error analysis for the discussed classes. The ResNet18 model exhibited higher false positive rates across all classes than the ViT model, indicating a tendency towards over-prediction. Conversely, the ViT model maintained lower false positives but demonstrated higher false negatives, particularly for Pedestrians and Cyclists, suggesting a cautious but potentially under-detecting approach. Notably, the ViT model was more balanced for the Motorcyclist class, with much lower false negatives than ResNet18, indicating more reliable detection.

\noindent
\textbf{Visualizations:} Figure~\ref{class-hm} shows heatmap visualization to illustrate the models' focus areas and any changes before the mitigation strategies. When trained with unbalanced classes, the tests show that both CNN models falsely classify motorcyclist and cyclist classes.

\begin{table}[!ht]
\centering
\caption{NDS and {\%IoU} for classes with ResNet18}
\label{tab:nds_data}
{\small
\begin{tabular}{|clc|c|c|}
\hline
\multicolumn{3}{|c|}{{\%IoU} of Classes}                                          & \multirow{2}{*}{NDS} & \multirow{2}{*}{\begin{tabular}[c]{@{}c@{}}Data \\ Condition\end{tabular}} \\ \cline{1-3}
\multicolumn{1}{|c|}{Pedestrian} & \multicolumn{1}{l|}{Cyclist} & Motorcyclist &                      &                                                                          \\ \hline
\multicolumn{1}{|c|}{86.1}       & \multicolumn{1}{l|}{53.6}    & 76.4         & 72.0                 & Normal                                                                     \\
\multicolumn{1}{|c|}{78.5}       & \multicolumn{1}{l|}{8.3}     & 40.7         & 42.5                 & Night                                                                      \\
\multicolumn{1}{|c|}{67.3}       & \multicolumn{1}{l|}{7.5}     & 18.9         & 31.2                 & Weather               \\
\multicolumn{1}{|c|}{85.7}       & \multicolumn{1}{l|}{52.8}    & 76.1         & 71.5                 &  Rotated                \\
\multicolumn{1}{|c|}{72.1}       & \multicolumn{1}{l|}{12.4}    & 21.4         & 35.3                 & Mixed   \\ \hline
\end{tabular}
}
\end{table}

\subsection{Discussion of Results}
For measuring IoU, mAP and NDS scores mentioned in Tables 1 and 2, we used a subset approach, where the representation of one class (e.g., pedestrian) is kept at 67{\%} and the other two classes have equal representation. This ratio is adjusted during iterations until all classes have the same representation. The purpose is to capture metrics in a scenario where the weighted class functions are normalized. Table~\ref{tab:nds_data} shows the Intersection over Union ({\%}IoU) for different classes and the NuScenes Detection Score (NDS) across various conditions for ResNet18. Pedestrian detection remains relatively high in all conditions, with {\%}IoU above 70{\%} except when combined adverse conditions are present. Cyclist detection suffers significantly in all but normal conditions, with {\%}IoU dropping to as low as 7.5{\%} during adverse weather. Motorcyclist detection is also affected by these conditions but to a lesser extent. The overall NDS reflects these trends, with the highest score of 72.0 in normal conditions and the lowest of 31.2 during adverse weather. The rotated condition only slightly impacts the {\%}IoU for pedestrians and motorcyclists but more so for cyclists. Mixed conditions present a scenario with a noticeable decrease in {\%}IoU across all classes and a resultant NDS of 35.3.

\begin{table}[!ht]
\centering
\caption{Mean Average Precision (mAP) of different classes With SqueezeNet under various conditions.}
\label{tab:map_data}
{\small
\begin{tabular}{|l|l|l|l|c|}
    \hline
    Pedestrian & Cyclist & Motorcyclist & Total & Data Condition \\ \hline
    86.3 & 54.7 & 77.8 & 72.93 & Normal \\
    78.1 & 7.5 & 38.3 & 41.3 & Night \\
    67.0 & 8.3 & 19.7 & 31.6 & Weather \\
    85.9 & 52.3 & 75.2 & 71.1 & Rotated \\
    73.2 & 14.8 & 22.1 & 36.7 & Mixed  \\ \hline
\end{tabular}
}
\end{table}

Table~\ref{tab:map_data} shows the mean average precision (mAP) of pedestrian, cyclist, and motorcyclist detection under various conditions. Under normal conditions, pedestrian detection is high at 86.3{\%}, while cyclist detection lags at 54.7{\%}. Motorcyclist detection is reasonably high at 77.8{\%}, leading to an overall mAP of 72.93{\%}. However, performance drops majorly at night and during adverse weather, with the lowest cyclist mAP at 7.5{\%} and 8.3{\%}, respectively. Interestingly, rotating images in normal conditions slightly reduces pedestrian and motorcyclist detection by about 1{\%}, but notably impacts cyclist detection, dropping to 52.3{\%}. The combination of all adverse conditions drastically reduces overall performance to a mAP of 36.7{\%}, showcasing the challenges faced by the models in less-than-ideal scenarios.

\begin{table}[!ht]
    \centering
    \caption{Class-wise Performance Metrics on the CNN models}
    \label{tab:class-wise-metrics}
    {\small
        \begin{tabular}{|l|l|l|l|l|l|}
            \hline
            Class        & Model    & \%IoU & NDS   & Sens. & Sel. \\ \hline
            Pedestrian   & ResNet18 & 75.3 & 82.1\% & 0.78       & 0.80       \\ 
            Cyclist      & ResNet18 & 68.4 & 79.5\% & 0.65       & 0.67       \\ 
            Motorcyclist & ResNet18 & 70.2 & 80.3\% & 0.70       & 0.72       \\ 
            Pedestrian   & SqueezeNet      & 78.1 & 85.4\% & 0.81       & 0.83       \\ 
            Cyclist      & SqueezeNet      & 72.6 & 82.7\% & 0.73       & 0.75       \\ 
            Motorcyclist & SqueezeNet      & 74.0 & 83.2\% & 0.76       & 0.77       \\ \hline
            \multicolumn{5}{l}{Sens. = Sensitivity Score, Sel = Selectivity Score}
        \end{tabular}
        }
    \end{table}

\begin{table}[!ht]
        \centering
        \caption{Baseline vs Post-Mitigation Performance}
         \label{tab:baseline-vs-post-mitigation}
         {\small
        \begin{tabular}{|l|l|l|l|}
            \hline
            Model        & Metric (Avg)    & Baseline &  Post Mitigation \\ \hline
            ResNet18     &  \%IoU  & 71.3\%    & 75.6\%          \\ 
            ViT          &  \%IoU  & 74.9\%    & 79.2\%          \\ 
            ResNet18     &  NDS   & 80.6\%    & 83.7\%          \\ 
            ViT          &  NDS   & 83.8\%    & 87.1\%          \\ \hline
        \end{tabular}
        }
\end{table}

\begin{figure*}[!ht]
\centering
\includegraphics[scale=0.45]{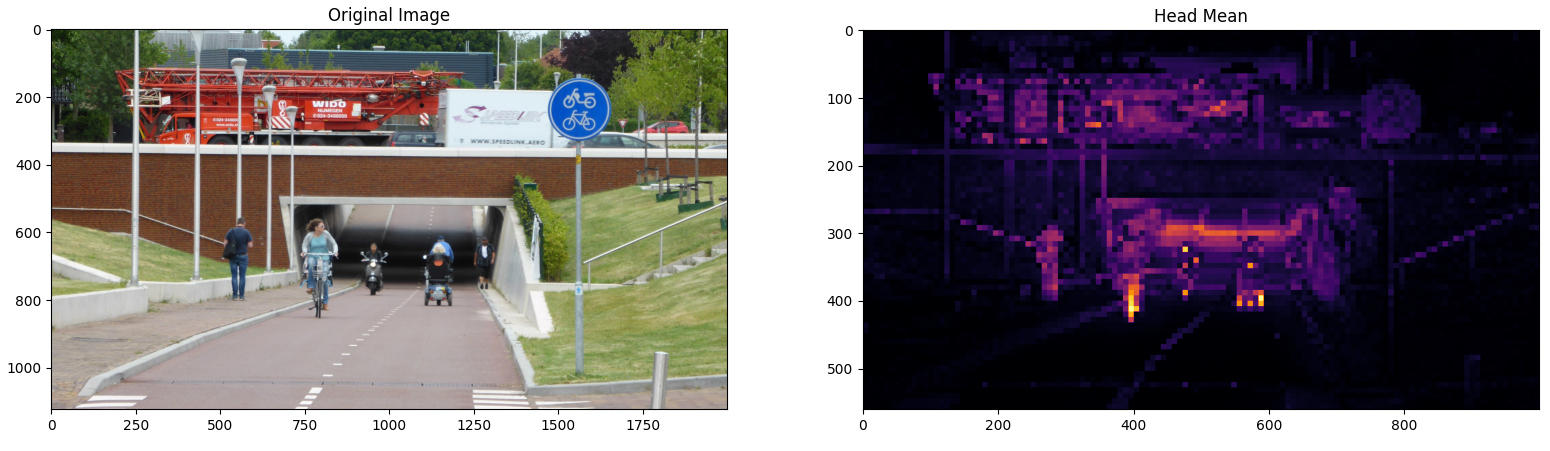}
\caption{Figure showing mean for all three classes}
\label{attn-mean}
\end{figure*}

\noindent
\textbf{Class-wise Performance Metrics:} Table~\ref{tab:class-wise-metrics} shows class-wise performance metrics for two models, ResNet18 and SqueezeNet. For the pedestrian class, SqueezeNet outperforms ResNet18 with a higher {\%}IoU and NDS. Cyclists and motorcyclists also see better {\%}IoU and NDS scores with SqueezeNet. Sensitivity and selectivity scores, which measure the models' ability to identify and differentiate classes, are consistently higher for SqueezeNet across all classes, indicating a more refined recognition capability. These metrics show overall better performance from SqueezeNet models in distinguishing and correctly identifying the classes.

\begin{figure} [!ht]
\centering
\includegraphics[scale=0.25]{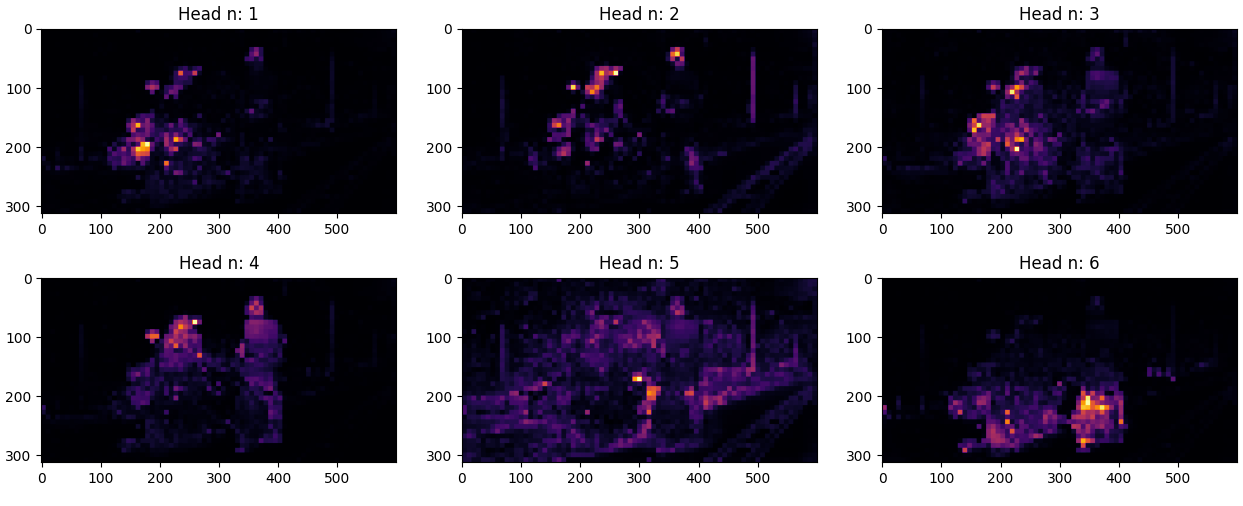}
\caption{Figure with head wise attention maps for motorcyclist class.}
\label{attn-motor}
\end{figure}

\textbf{Baseline vs Post-Mitigation Performance:} Table~\ref{tab:baseline-vs-post-mitigation} shows average Intersection over Union (IoU) and NuScenes Detection Score (NDS) before and after applying bias mitigation techniques for ResNet18 and Vision Transformer (ViT). Both models show an improvement in {\%}IoU and NDS after mitigation, with ViT exhibiting a larger increase in NDS. ResNet18's IoU improved by 4.3{\%}, and its NDS by 3.1{\%}, while ViT's IoU and NDS improved by 4.3{\%} and 3.3{\%}, respectively. This demonstrates the effectiveness of the bias mitigation strategies.

\begin{figure}[!ht]
\centering
\includegraphics[scale=0.30]{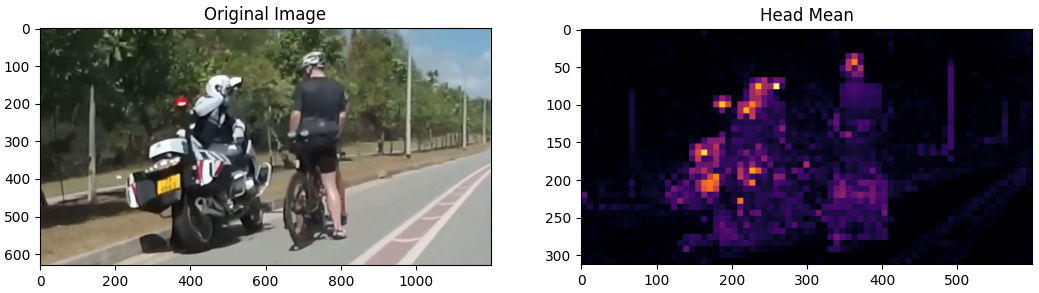}
\caption{Figure showing mean for motorcyclist classes.\\}
\label{mean-motor}
\end{figure}

\begin{figure}[!ht]
\centering
\includegraphics[scale=0.18]{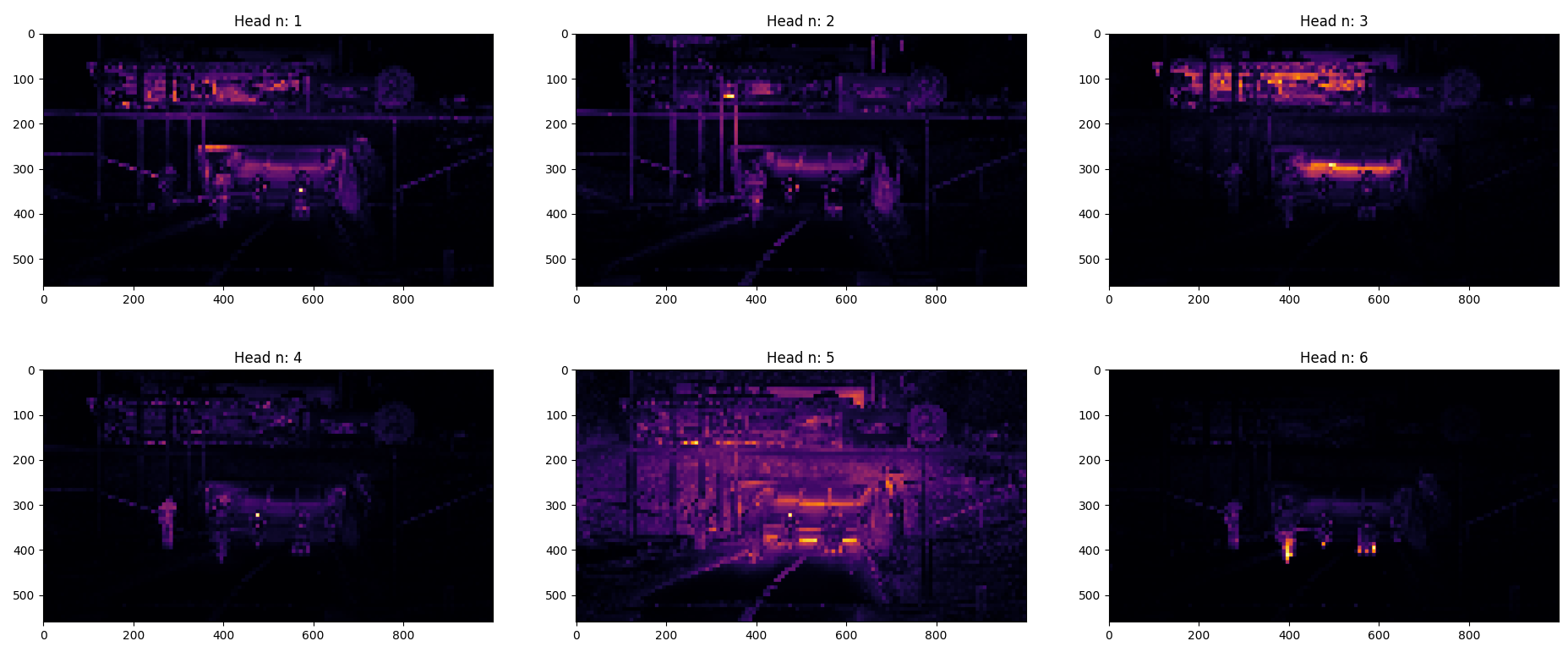}
\caption{Figure showing head-wise attention weights for all classes.\\}
\label{attn-cls}
\end{figure}

\textbf{Visualizations:} Figure~\ref{mean-motor} and Figure~\ref{attn-motor} shows mean and attention maps for the motorcyclist class by still capturing features from the cyclist present in the input image. The model used for testing the images has dominant weights from the motorcyclist class, thus fails to capture the features of cyclist. Similarly, Figure~\ref{attn-mean}, and Figure~\ref{attn-cls} shows the results from the test which has presence of equal number of classes, but as it can be seen from the image, the model also assign weights to features represented by cyclist and pedestrians to other classes. In a summary the key takeways can be described as:

\begin{itemize}
    \item For CNN models, error analysis shows ResNet18 has higher false positive rates than SqueezeNet.
    \item Post-mitigation performance, which involves resampling approaches, shows improvements in both CNN (ResNet18) and ViT.
    \item Under all conditions, the pedestrian class had higher detection. However, overall performance metrics show a noticeable decrease, seeking robust model development.
    \item Equal number of samples from classes in a subset, does not necessarily ensure an equal representation of behaviour metrics during model learning. 
\end{itemize}

\section{Conclusion and Future Work}
This study explored the critical issue of class imbalance in driving dataset, specifically targeting the under-represented classes and its impact on the performance of AI models. Through a comprehensive methodology encompassing dataset analysis, model evaluation, and bias impact assessment, we detect disparities in the representation and detection of vulnerable road users. Our tests with popular CNN models and Vision Transformers have revealed how dataset biases can lead to alternate learning outcomes, adversely affecting the accuracy and correctness of perception systems. Further, we have implemented and evaluated bias mitigation techniques, which include cost-sensitive learning and targeted data augmentation. These methods have shown promising results in improving model performance, especially in IoU and NDS metrics. Furthermore, introducing a dynamic framework for ongoing bias assessment and model recalibration signifies an important step towards developing more equitable AI systems in autonomous driving.


\section*{Acknowledgment}
The authors gratefully acknowledge funding from European Union's Horizon 2020 Research and Innovation programme under the Marie Sk\l{}odowska Curie grant agreement No. 956090 (APROPOS), SPATIAL project under grant agreement No~101021808, and CONCORDIA project under grant agreement No~830927.

\ifCLASSOPTIONcaptionsoff
  \newpage
\fi

\bibliographystyle{plain}
\bibliography{bare_jrnl.bib}

\end{document}